\documentclass{article}
\usepackage{amsmath,epsfig}
\usepackage[preprint]{spconfa4}

\usepackage{graphicx}
\usepackage{floatrow}
\usepackage{amsmath}
\newfloatcommand{capbtabbox}{table}[][\FBwidth]

\let\OLDthebibliography\thebibliography
\renewcommand\thebibliography[1]{
  \OLDthebibliography{#1}
  \setlength{\parskip}{0pt}
  \setlength{\itemsep}{0pt plus 0.3ex}
}

\begin{document}\sloppy

\def\x{{\mathbf x}}
\def\L{{\cal L}}

\title{Joint learning of object graph and relation graph for\\ visual question answering}

\name{Hao Li$^{1, \dagger}$, Xu Li$^{2}$, Belhal Karimi$^{2}$, Jie Chen$^{1,3}$, Mingming Sun$^{2, \ast}$\thanks{\noindent $^{\ast}$Corresponding Author}
\thanks{$^{\dagger}$ Work was done during first author’s internship at Baidu.}
\thanks{This work is supported in part by the Nature Science Foundation of China (No.61972217, No.62081360152), Natural Science Foundation of Guangdong Province in China (No.2019B1515120049, 2020B1111340056).}}
\vspace{-0.1in}
\address{$^{1}$School of Electronic and Computer Engineering, Peking University, China;\\ $^{2}$Cognitive Computing Lab, Baidu Research, China; $^{3}$ Peng Cheng Laboratory, Shenzhen, China;\\ lihao1984@pku.edu.cn, chenj@pcl.ac.cn, \{lixu13, belhalkarimi, sunmingming01\}@baidu.com}

\maketitle

\begin{abstract}
Modeling visual question answering~(VQA) through scene graphs can significantly improve the reasoning accuracy and interpretability. However, existing models answer poorly for complex reasoning questions with attributes or relations, which causes \textbf{false attribute selection} or \textbf{missing relation} in Figure~1(a). 
It is because these models cannot balance all kinds of information in scene graphs, neglecting relation and attribute information.
In this paper, we introduce a novel Dual Message-passing enhanced Graph Neural Network (DM-GNN), which can obtain a balanced representation by properly encoding multi-scale scene graph information.
Specifically, we (i)~transform the scene graph into two graphs with diversified focuses on objects and relations; Then we design a \textit{dual structure} to encode them, which increases the weights from relations (ii)~fuse the encoder output with attribute features, which increases the weights from attributes; (iii)~propose a \textit{message-passing mechanism} to enhance the information transfer between objects, relations and attributes.
We conduct extensive experiments on datasets including GQA, VG, motif-VG and achieve new state of the art.
\end{abstract}
\section{Introduction}
\label{sec:intro}

VQA tasks require a model to answer a free-form natural language question using visual information from an image. Scene graph (SG) reasoning is an essential instance of VQA tasks~\cite{DBLP:journals/corr/abs-2007-01072}. 
The model extracts objects' names, attributes, and relations from the input images and organizes them into a graph representation to generate the scene graph.

SG representation modeling displays several virtues over classical VQA techniques since the features in SG are presented in plain and free text form~\cite{DBLP:journals/corr/abs-2101-05479} and the graph structures of SG have better interpretability~\cite{DBLP:conf/bmvc/ZhangCX19}.
In this contribution, two reasoning methods on scene graphs are proposed: (i) consider scene graphs as probabilistic graphs and iteratively update nodes' probabilities using soft instructions extracted from questions~\cite{DBLP:conf/nips/HudsonM19}; (ii) apply Graph Neural Network (GNN) into scene graphs~\cite{inproceedings-fstt,DBLP:conf/iccv/LiGCL19} to learn joint representations of nodes and their relations, and then feed these representations into a predictor to get the answer. Scene graph reasoning frameworks are useful in VQA~\cite{yang2020prior}. However, there still remain imperfections dealing with complex reasoning questions.

\begin{figure}[t!]
    \centering 
    \includegraphics[scale=0.6]{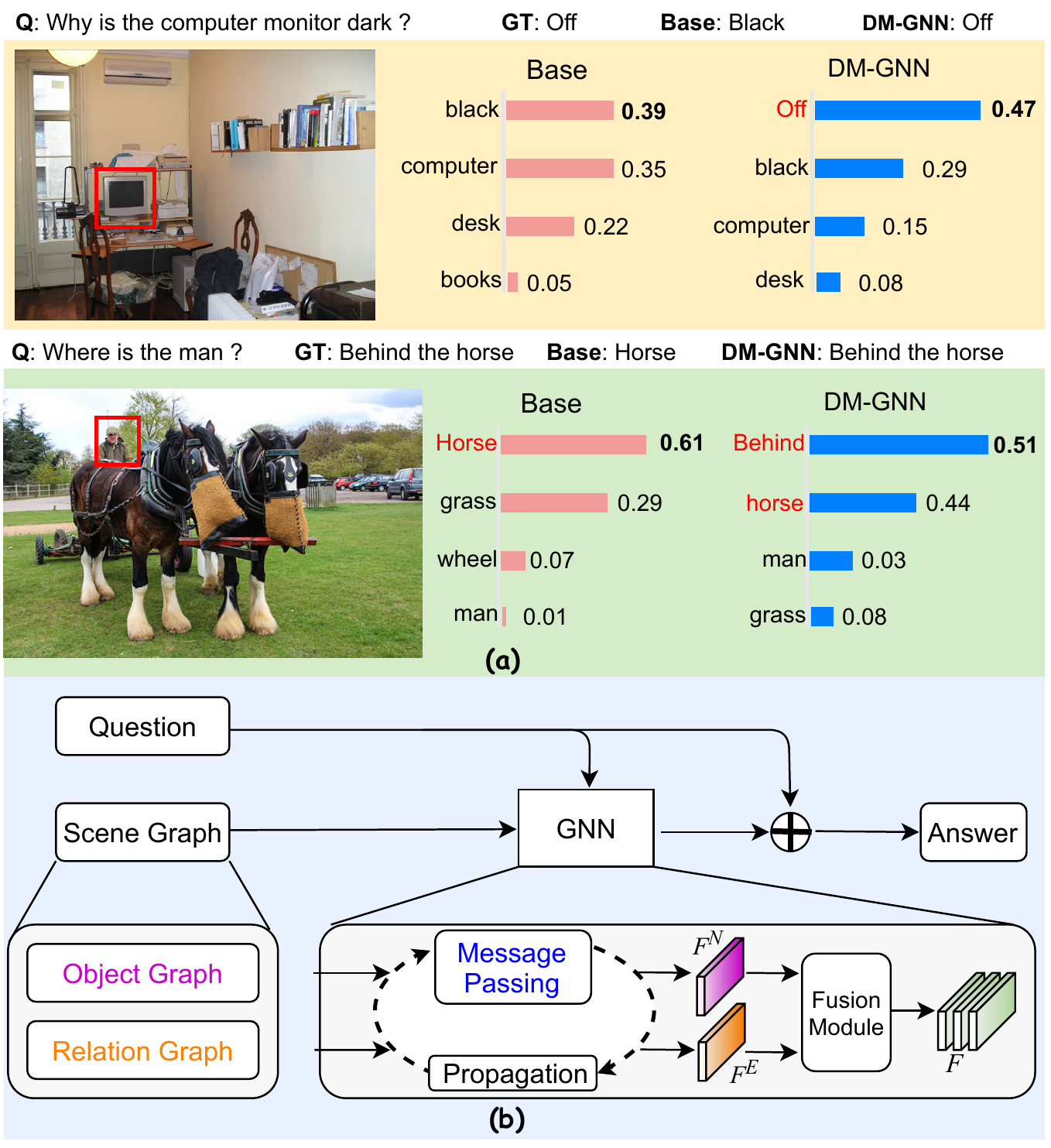} 
    \caption{(a) Two key issues of traditional scene graph based models that we address: \textbf{false attribute selection} (top: select attribute ``black" instead of attribute ``off") and \textbf{missing relation} (bottom: missing ``Behind") (b) Overview of our DM-GNN model. $F^N$ and $F^E$ are the object feature map and the relation feature map. $F$ is the full-scale feature map.} \label{scene-graph}
    \vspace{-0.27in}
\end{figure}

\begin{figure*}[ht] 
    \vspace{-0.3in}
    \centering 
    \includegraphics[width=1\textwidth]{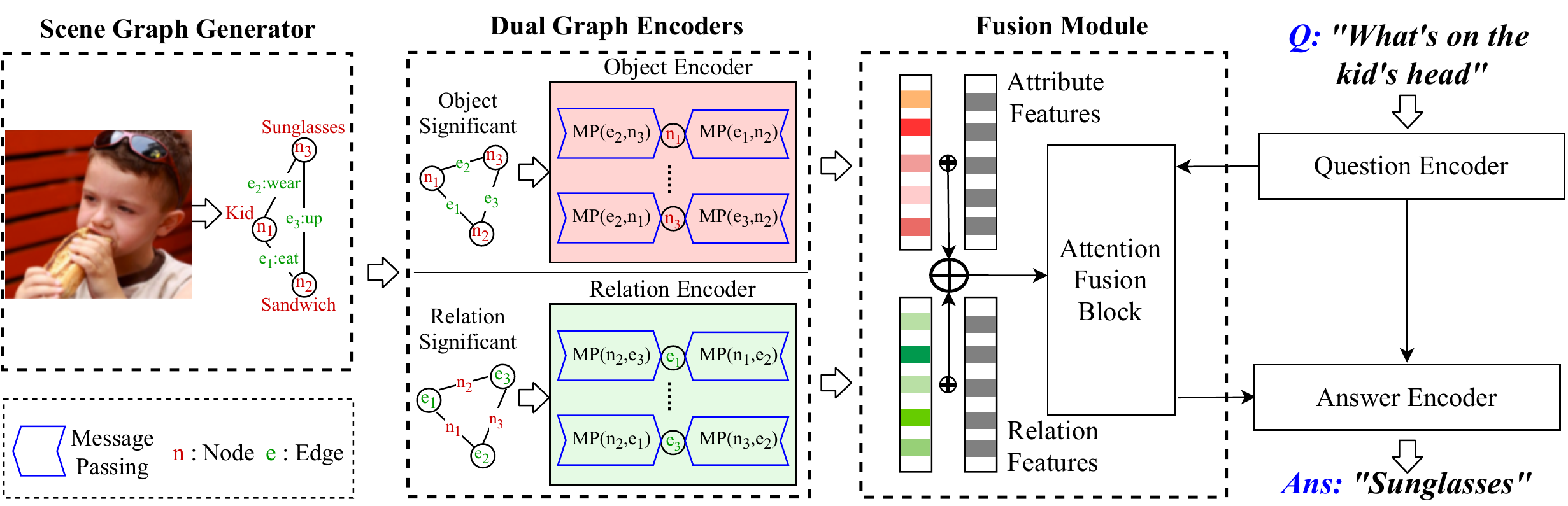} 
    \vspace{-0.3in}
    \caption{Model structure of the Dual Message-passing enhanced Graph Neural Networks. MP stands for the message-passing module. Images are transformed into scene graphs by the scene graph generator. The object-significant form and relation-significant form of the scene graph are injected into the object encoder and the relation encoder. Nodes' representations are generated from the sum of the MP modules. The representations are then fused with question representations to predict answers.} 
    \label{fig2} 
\end{figure*}

First, existing models tend to predict wrong answers for complex reasoning questions with attributes. 
Consider the ``why'' question in Fig.~\ref{scene-graph}(a) as an example, \textbf{false attribute selection} occurs because the model cannot associate ``off'' relation with ``dark monitor'' object. The attribute selections require comprehensive supervision from objects, relations, and attributes, but existing methods focus too much on objects while ignoring attributes. Generally, information from objects and relations connected to them are reconstructed into object features in GNN-based methods~\cite{xu2019spatial}. However, these encoding methods lack information from objects' attributes. 
NSM methods~\cite{DBLP:conf/nips/HudsonM19} use soft instructions to update answer possibilities, but they treat attributes as secondary information.

Second, existing approaches answer poorly for complex questions that require information about relations. For instance, for localization questions, as in the ``where'' type of questions, in Fig.~\ref{scene-graph}(a), we observe that \textbf{missing relation} occurs because the model cannot capture the relation information ``Behind''. Existing models have a strong bias towards object features, while considering relation as references. The unbalanced focus on objects and relations makes the models fail to learn discriminative representations for relations.

To improve the balance of all kinds of information in scene grpahs, we propose the Dual Message-passing enhanced Graph Neural Network (DM-GNN) for VQA, introducing a novel scene graph reasoning model that extracts balanced feature maps from objects, attributes, and relations information in scene graphs. 
Concretely, as shown in Fig.~\ref{scene-graph}(b), our DM-GNN model is composed of a scene graph generator, a question encoder, dual graph encoders, and a fusion module. 
Besides, to balance the importance of objects and relations, we transform scene graphs into a relation-significant modality, where nodes represent relations and edges represent objects, and an object-significant modality, in which nodes represent objects and edges represent relations. 
After receiving scene graphs in two modalities, dual graph encoders can produce feature maps focusing on relations and objects.

Furthermore, to enhance the information transfer between objects, relations and attributes, we modify the gated graph neural network (GGNN) structure in our DM-GNN by adding the message-passing module. It is a bidirectional GRU that guides the internal information flow. The encoder captures information from nodes, edges, and adjacent nodes that connect to them. 
In Fig.~\ref{scene-graph}(b), the output feature map of the encoder passes through the fusion module, where the attribute features are explicitly modeled into the feature map to increase the information weight from attributes. Then the feature map passes through multi-head attention layers using question features extracted from the question encoder. Hence, the model dynamically focuses on the critical parts of the questions and uses the most similar part of the scene graph as the most adequate answer. Our main contributions are as follows:\vspace{-0.07in}
\begin{itemize}
\setlength{\itemsep}{5pt}
\setlength{\parsep}{5pt}
\setlength{\parskip}{5pt}
\item We analyse that existing models answer imperfectly for complex reasoning questions with attributes or relations due to the unbalance focus on three information types in scene graphs, which contain objects, relations and attributes.\vspace{-0.06in}


\item We propose a novel DM-GNN model containing a dual encoder structure and a message-passing module. Our model can obtain a balanced representation by properly encoding multi-scale scene graph information.\vspace{-0.06in}

\item Experimental results on various datasets show that DM-GNN effectively improves the reasoning accuracy on semantically complicated questions.
\end{itemize}
\vspace{-0.1in}

\section{Related Work}

\noindent\textbf{Visual Question Answering.}
Most VQA approaches use sequential models~\cite{DBLP:conf/naacl/DevlinCLT19} to encode questions and CNN-based pretrained models~\cite{DBLP:conf/cvpr/PatroN18} to encode images. Then they use attention methods~\cite{DBLP:conf/nips/LuYBP16,DBLP:conf/iclr/HudsonM18} to fuse features from images and questions.
Transformer models~\cite{DBLP:conf/aaai/LiDFGJ20} achieve outstanding performances on VQA tasks, yet they are heavy to train and hard to explain~\cite{DBLP:journals/corr/abs-2106-07139}. Instead, the scene graph model stands for an alternative that is more lightly and explainable.

\vspace{0.05in}
\noindent\textbf{Scene Graph Generation and Reasoning.}
Scene graph generation (SGG) methods~\cite{DBLP:conf/cvpr/TangNHSZ20} use object detection methods to extract region proposals from images. Scene graph can promote explainable reasoning for downstream multimodal tasks such as VQA~\cite{DBLP:conf/bmvc/ZhangCX19}. 
In typical scene graph reasoning models, NSM~\cite{DBLP:conf/nips/HudsonM19} 
performs sequential reasoning over the scene graph by iteratively traversing its nodes.
Other models~\cite{inproceedings-fstt, DBLP:conf/iccv/LiGCL19,DBLP:journals/corr/GraphVQA} use GGNN~\cite{DBLP:journals/corr/LiTBZ15} based model to encode scene graphs. However, previous works are hard to fully utilize the attribute information and learn the comprehensive representation of SG.

\vspace{0.05in}
\noindent\textbf{Graph Neural Network.}
GNN~\cite{DBLP:conf/iclr/VelickovicCCRLB18} is designed to infer on data described by graphs. 
\cite{DBLP:conf/cncl/WangGCL16,DBLP:conf/aaai/WangCGL18} apply GNN-based models on knowledge graphs, which are similar to scene graphs.
However, existing GNN-based models cannot effectively process graphs with node attributes and complicated labels.
Our DM-GNN model can learn a comprehensive and balanced representation using full-scale scene graph information from objects, attributes, and relations to overcome these problems.

\section{DM-GNN Methodology}

Our proposed architecture is illustrated in Fig.~\ref{fig2}. 
We use the scene graph generator from~\cite{tang2020sggcode}. In the question encoder, semantic questions are first projected into an embedding space using GLOVE pretrained word embedding model~\cite{pennington-etal-2014-glove}. 
Then we use long short-term memory (LSTM) networks to generate questions representation $q \in R^{dim}$, where dim is the dimension of the question representation.
We introduce our dual graph encoders and fusion module in following subsections.

\vspace{-0.07in}
\subsection{Object/Relation-Significant Graph}
We organize scene graphs into object-significant graphs and relation-significant graphs. 

\medskip
\textbf{Object-Significant Graph.} We define the object significant graph as $\emph{G}_{obj}$, where each node represents an object in the image and each edge represents a relation between two objects. Define $\emph{N}$ as the node set and $\emph{E}$ as the edge set. For $n_i, n_j \in \emph{N}$, $e_k \in \emph{E}$, $<n_i$ - $e_k$ - $n_j>$ denotes the relation tuple that represents relation $e_k$ from object $n_i$ to object $n_j$.

\medskip
\textbf{Relation-Significant Graph.} We define relation significant modality as $\emph{G}_{rel}$, where each node represents a relation between objects in the image and each edge represents an object, which is completely opposed to the object-significant modality. For $e_i, e_j \in \emph{E}$, $n_k \in \emph{N}$, $<e_i - n_k - e_j>$ represents the relations $e_i$ and $e_j$ have a shared object $n_k$.

\begin{figure}[t] 
\vspace{-0.1in}
    \includegraphics[width=1\textwidth]{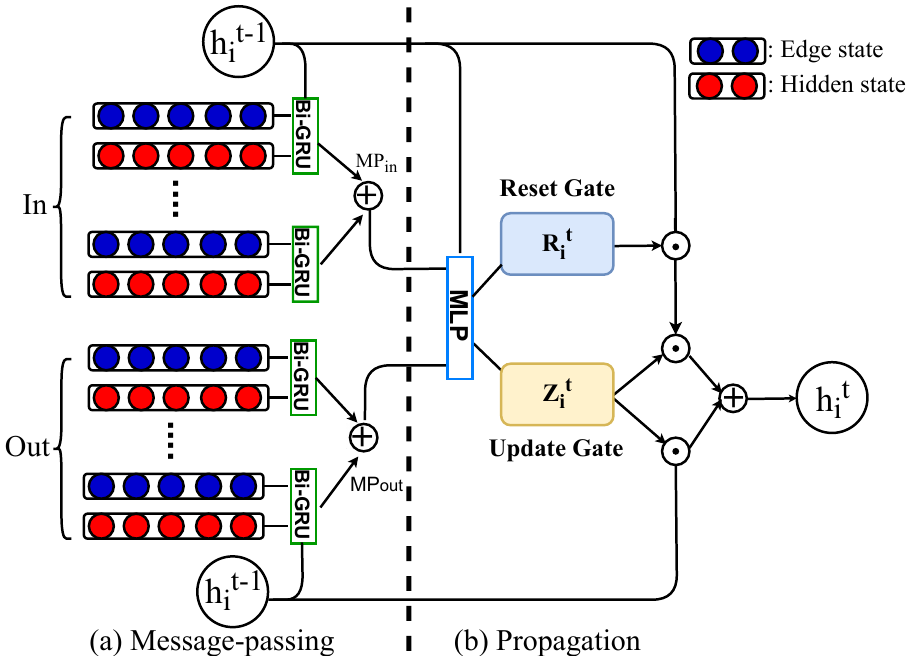} 
    \caption{Overview of the pipeline at timestep t. (a) The message-passing module generates $MP_{in}$ and $MP_{out}$ as incident and output information gains. (b) The propagator uses information gains and $h_i^{t-1}$ to update $h_i^t$.}
    \label{EF} 
\end{figure}

\medskip
\textbf{Attribute types.} Define \emph{L} as attribute types. 
For each node $n_i\in\emph{N}$, we define a set of $\emph{L}+1$ property variables ${\left\{ n_i^l\right\}}_{l=0}^L$, where $n_i^{0}$ represents $n_i$'s name embedding and $n_i^l$ represents the embedding of node $n_i$'s $l^{th}$ attribute.

\subsection{Dual Encoders}
We apply two GGNN-based encoders for object significant graph and relation significant graph. 
The encoder for object graph focuses on object features and the encoder for relation graph focuses on relation features. 
The dual structure can balance the importance of relations and objects.

Prior to encoding, every input scene graph is transformed into an information tuple ($\emph{N}, \emph{E}, \emph{A}_{in}, \emph{A}_{out}$): \emph{N} and \emph{E} are collections of node embeddings and edge embeddings. $\emph{A}_{in}$ and $\emph{A}_{out}$ are the adjacency matrix of incident and output edges.


Let $h_i^t$ is the hidden state of node $n_i$ in the encoder at timestep $\emph{t}$, then at $\emph{t}=0$, we initialize $h_i^0$ as the GLOVE embedding of $n_i$ with zero padding:


\textbf{Message-passing Module.}
To enhance the information transfer from edges and adjacent nodes to the updating nodes, we use the message-passing module (MP) in Fig.~\ref{EF}(a).
MP module comes as a replacement of the fully-connected layers from the original GNN model. 
Consider a tuple $<{n}_i, {e}_k, {n}_j>$ as the processing sample of the message-passing module. 
The embedding state, noted $\textit{e}_k$, of the edge ${e}_k$ and neighbor node ${n}_j$'s hidden state $\textit{h}_j$ are injected into a bidirectional GRU network as input sequence while the node ${n}_i$'s hidden state $\textit{h}_i$ is injected as the GRU's initial hidden state. 
The output of the GRU represents the updating information for hidden state $\textit{h}_i$, which corresponds to the key information from edge ${e}_k$ and node ${n}_j$ that is related to node ${n}_i$. 
The sum of every GRU output is ${n}_i$'s total information gain from ${n}_i$'s adjacent nodes and edges. 
We detail the message-passing module formula as follows:
\begin{gather}
    MP_i(A_{in}) = \sum\limits_{k,j}^{<n_i,e_k,n_j>\in A_{in}} \text{GRU}([\textit{e}_k, \textit{h}_j], \textit{h}_i) \, , \\ 
    MP_i(A_{out}) = \sum\limits_{k,j}^{<{n}_j,{e}_k,{n}_i>\in A_{out}} \text{GRU}([\textit{e}_k, \textit{h}_j], \textit{h}_i) \, ,
\end{gather}
where $MP_i(A_{in})$ is ${n}_i$'s incident information gain, and $MP_i(A_{out})$ is ${n}_i$'s output information gain.

\textbf{Propagation Module. } In Fig.~\ref{EF}(b), at timestep $t$, the hidden states of all nodes are updated by the following gated propagator module:
\begin{gather}
    k_{i}^t = [MP_i^t(A_{in}), MP_i^t(A_{out})] \, ,
\end{gather}
where $k_{i}^t$ is the node ${n}_i$'s representation from all its incident edges, output edges and adjacent nodes.

Then, we incorporate information from adjacent nodes and from the previous timestep leading to an update of each node's hidden state:
\begin{gather}
    c_i^t = [h_i^{(t-1)}, k_{i}^{(t-1)}] W + b \, ,\\
    z_i^t = \sigma(U^z c_i^t) \, , \\
    r_i^t = \sigma(U^r c_i^t)\, , 
\end{gather}
where $W, U^z$ and $U^r$ are referred to as the trainable weight matrices. At timestep $t$, $z_i^t$ and $r_i^t$ are the update and reset gates, respectively. Then we have:
\begin{gather}
    \tilde{h}_i^t = \text{tanh}(U_1 k_{i}^{(t-1)} + U_2(r_i^t \odot h_i^{(t-1)})) \, ,\\
    h_i^t = (1-z_i^t)\odot h_i^{(t-1)} + z_i^t\odot \tilde{h}_i^t \, .
\end{gather}
Here, $U_1$ and $U_2$ denote the trainable parameters of the linear layers, the operator $\odot$ is the element-wise multiplication. $\sigma$ is the ReLU function.
After $T$ steps, the encoder generates the final hidden state map $G$ of the graph. 
Finally, we compute the graph embedding $g_i \in G$ for node ${n}_i$ as follows:
\begin{equation}
    g_i = \sigma( \emph{f}(h_i^T, n_i)) \, ,
\end{equation}
where $\emph{f}(h_i^T, n_i)$ is multi-layer perceptron (MLP) which receives the concatenation of $h_i^T$ and $n_i$.

\subsection{Fusion Module and Answer Predictor}

Once the dual encoders, embedded in our model, output the node and relation features, we first fuse the attributes into feature maps.
For node feature map $G^N$ and relation feature map $G^E$, the fusion feature map $F^N$ and $F^E$ are defined as

{\small \begin{equation}
F_i^N= \begin{cases}
[g_i^N, n_i^0]  \\
\cdots, \\
[g_i^N, n_i^L]
\end{cases} 
,\, F_j^E = [g_j^E, e_j] , \,
    F = [F^N, F^E]  ,
\end{equation}
}

where $F_i^N$ indicates the fusion features of node $i$ and $g_i^N$ is node $i$'s representation from the encoder. $n_i^l$ is the attribute embedding of node $i$. 
$F_j^E$ corresponds to the fusion feature of edge $j$. $g_j^E$ is edge $j$'s representation from the encoder. 
$e_j$ is $j$-th edge original embedding. 
The full-scale feature map, noted $F$, is obtained by concatenating $F^N$ and $F^E$.

Then, the question embedding $\emph{q}$ and the full-scale feature map $F$ are fed into a multi-head attention layer.
The reasoning vector, noted $\emph{r}$, and which stems from the graph and the question, is computed using a weighted sum of the feature map using the scores output from the attention layer,
\begin{gather}
    r = \text{Attention}(F, q) \, . 
\end{gather}

In answer predictor module, we adopt a two-layer MLP noted by $f(\cdot)$. 
This MLP can be viewed as a classifier over the set of candidate answers. 
The input of the answer predictor is the concatenation vector $(\emph{q},\emph{r})$. 
Such a classifier has been applied in many VQA models~\cite{DBLP:conf/nips/HudsonM19,DBLP:conf/nips/LuYBP16}.
The answer $\hat{a}$ reads:
\begin{gather}
    \hat{a} = \mathop{\arg\max}(\text{softmax}(f((\emph{q},\emph{r}))))\, .
\end{gather}


\vspace{-0.1in}
\section{Experiments}\label{sec:experiments}

\begin{table*}[ht]
\resizebox{0.9\textwidth}{!}{
\centering
    \begin{tabular}{l|lllll|lllll}
     \hline
     \textbf{Dataset}&\multicolumn{5}{c|}{VG-GroundTruth}&\multicolumn{5}{c}{Motif-VG}\\
     \hline
     Question type&\textbf{What}&\textbf{Where}&\textbf{Who}&\textbf{Why}&\textbf{Overall}&\textbf{What}&\textbf{Where}&\textbf{Who}&\textbf{Why}&\textbf{Overall}\\
     \hline
     Percentage &(54\%) &(17\%) &(5\%) &(3\%) &(100\%)&(54\%) &(17\%) &(5\%) &(3\%) &(100\%)\\
     \hline
     NSM~\cite{DBLP:conf/nips/HudsonM19} &33.1 &51.0 &49.8  &12.3 &45.1 &31.8 &53.1 &47.6 &10.9 &43.1\\
     F-GN~\cite{DBLP:conf/bmvc/ZhangCX19}&60.9 &62.0 &63.3 &50.9 &60.1 &58.7 &60.4 &61.8 &49.0 &60.0\\
     U-GN~\cite{DBLP:conf/bmvc/ZhangCX19}&61.6 &62.4 &63.9 &50.3 &60.5  &59.4 &60.3 &66.6 &48.1 &60.5\\
     FSTT~\cite{inproceedings-fstt} &65.5 &70.1 &68.3 &91.5 &65.6 &48.8 &49.2 &40.6 &70.3 &48.1\\
     ReGAT~\cite{DBLP:conf/iccv/LiGCL19} &72.1 &64.4 &72.7 &92.3 &71.2 &75.4 &57.6 &69.1 &91.8 &69.9\\
     DM-GNN (ours) &\textbf{75.9} &\textbf{73.1} &\textbf{82.6} &\textbf{98.8} &\textbf{75.4} &\textbf{79.4} &\textbf{62.7} &\textbf{72.8} &\textbf{96.1} &\textbf{72.9}\\
     \hline
    \end{tabular}}
    \vspace{-0.2in}
    \caption{
    \label{VG-detail}
Performance on different question types of VG dataset.}
\end{table*}

\begin{table*}[htbp]
\resizebox{0.93\textwidth}{!}{
    \begin{floatrow}
    \capbtabbox{
     \begin{tabular}{llllll}
       \hline
       \textbf{Models}&\textbf{Binary$\uparrow$}&\textbf{Open$\uparrow$}&\textbf{Validity$\uparrow$}&\textbf{Distribution$\downarrow$}&\textbf{Acc.$\uparrow$}\\
    \hline
     Human &91.20 &87.40 &98.90 &- &89.30\\
     BottomUp &66.64 &34.83 &96.18 &5.98 &49.74\\
     MAC &71.23 &38.91 &96.16 &5.34 &54.06\\
     SK T-Brain &77.42 &43.10 &96.26 &7.54 &59.19\\
     PVR &77.69 &43.01 &\textbf{96.45} &5.80 &59.27\\
     GRN &77.53 &43.35 &96.18 &6.06 &59.37\\
     Dream &77.84 &43.72 &96.38 &8.40 &59.72\\
     LXRT &77.76 &44.97 &96.30 &8.31 &60.34\\
     NSM &78.94 &49.25 &96.41 &\textbf{3.71} &63.17\\
     ReGAT &\textbf{83.57} &62.58 &92.70 &9.32 &70.50\\
     DM-GNN(ours) &69.79 &\textbf{72.21} &93.80 &3.78 &\textbf{71.21}\\
    \hline
    \end{tabular}
    }{
     \vspace{-0.2in}
     \caption{Performance on the GQA dataset.}
     \label{GQA}
    }
    \capbtabbox{
    \begin{tabular}{ll}
      \hline
      \textbf{Models}&\textbf{Acc.}\\
      \hline
      \textbf{Base} & 35.4 \\
      \hline
      \textbf{Base-Obj} & 35.4 \\
       ~+\emph{MP}& 39.3{$_{\textcolor{red}{(+3.9)}}$} \\
       ~+\emph{Dual} &67.9{$_{\textcolor{red}{(+32.7)}}$}\\
      \hline
      \textbf{Base-Rel} & 35.2 \\
       ~+\emph{MP}& 38.8{$_{\textcolor{red}{(+3.6)}}$} \\
       ~+\emph{Dual} &67.9{$_{\textcolor{red}{(+32.5)}}$}\\
      \hline
      \textbf{DM-GNN(ours)} & 75.2\\
       ~+ (w/o \emph{attr}) & 71.6{$_{\textcolor{blue}{(-3.6)}}$}\\
       ~+ (w/o \emph{rela}) & 74.5{$_{\textcolor{blue}{(-0.7)}}$} \\
       ~+ (w/o \emph{QF}) & 54.9{$_{\textcolor{blue}{(-20.3)}}$}\\
      \hline
    \end{tabular}
    }{\vspace{-0.2in}
     \caption{Ablation Study on VG.}
     \label{ablation-study}
    }
    \end{floatrow}
}
\end{table*}

\vspace{-0.1in}
\subsection{Empirical Results}

\quad \textbf{Results on VG dataset.} Table~\ref{VG-detail} reports the results on the test sets of the VG ground truth dataset and the motif-VG dataset. Compared to the baseline models, we can observe that our DM-GNN model outperforms the others at $\textbf{3\%}$-$\textbf{4\%}$. In addition, we provide detailed results on the VG dataset and motif-VG dataset with different question types. Compared to the other scene graph based VQA models, our model performs well in ``what", ``where", ``who" and ``why" types. On the VG dataset, our model has $\textbf{6.5\%}$ accuracy improvement in ``\textbf{why}" type questions, which highly requires VQA models' ability to jointly exploit objects, relations and attributes. 


\begin{figure*}[ht] 
    \vspace{-0.5in}
    \centering 
    \includegraphics[width=1\textwidth]{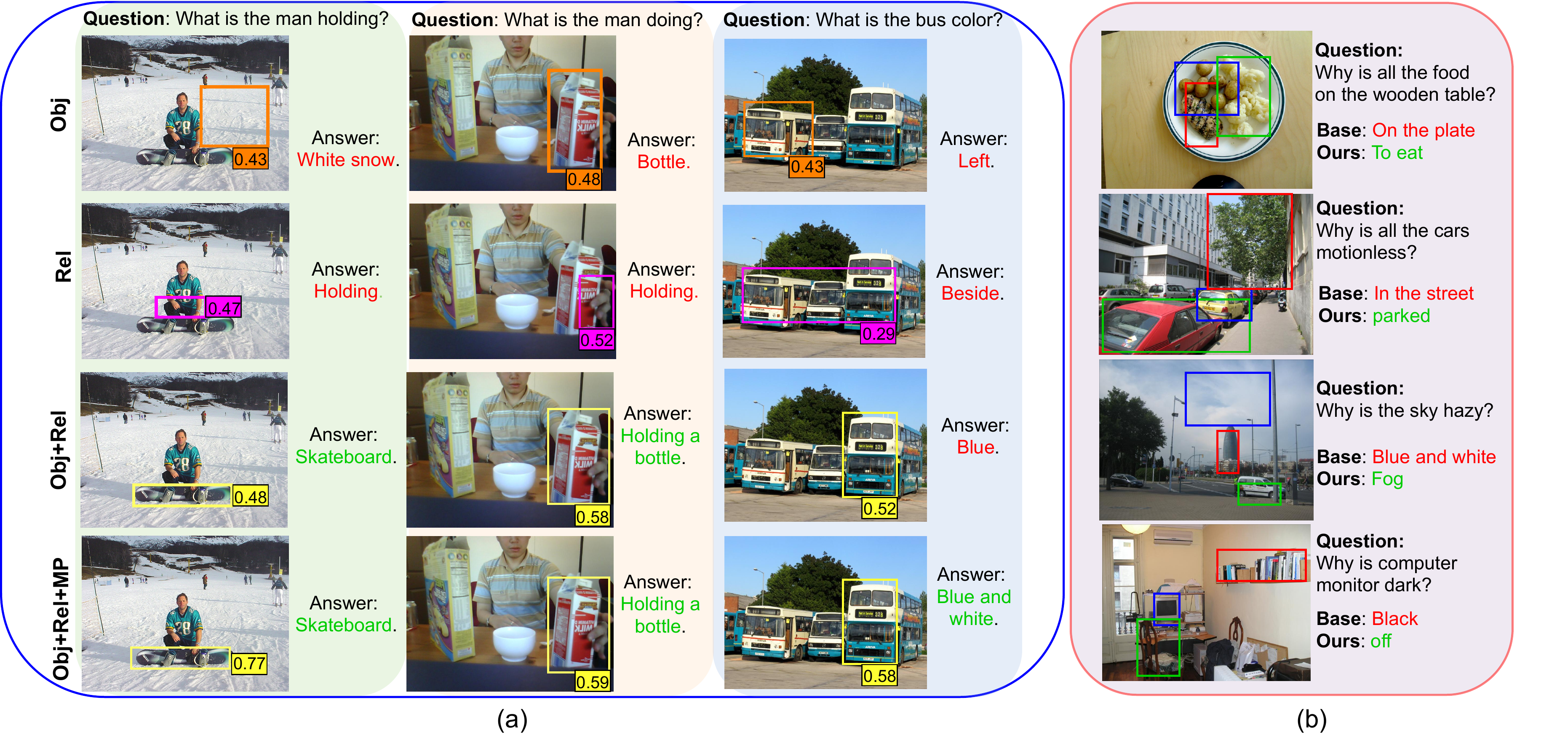} 
    \vspace{-0.2in}
    \caption{Examples of generated answers and top attention scores. Answers in red means wrong and green means right. (a) and (b) show the visualization of ``what" and ``why" question types. From (a), we observe that our approach can correctly select answer from objects, relations and attributes. From (b), we note that our model can handle comprehensive reasoning questions.} 
    \label{visual} 
    \vspace{-0.2in}
\end{figure*}


\textbf{Results on GQA dataset.} We report in Table~\ref{GQA} the detailed results on the test sets of the GQA dataset. Our DM-GNN model outperforms baselines. We also evaluate our model and other baselines across GQA dataset's various metrics, where ``Binary" and ``Open" stand for binary-answer and open domain questions. ``Distribution" corresponds to the distance between prediction distribution and standard answer distribution. For open domain questions which are difficult for reasoning, our model outperforms the others by $\textbf{9.63\%}$. For distribution metric, our model also achieves 2nd score. 
However, the “Binary” question is challenging for DM-GNN although it achieves SOTA performances. Specifically, the advantage of DM-GNN is to correctly locate key objects, relations and attributes. That is why it works well for the open domain~(``what, why, how'') questions. But for “yes/no” questions, whose answers are not explicit in scene graphs, DM-GNN is easy to locate but hard to correctly answer.




\vspace{-0.1in}
\subsection{Ablation Study}
We compare several ablated forms of DM-GNN with our complete model on the VG dataset. 
The accuracy for each variant of DM-GNN are reported in Table~\ref{ablation-study}. 
We use the raw GGNN network as the \emph{Base} model.
The \emph{Base-Obj} and \emph{Base-Rel} model represent the original GGNN network processing \emph{object-significant} graph and \emph{relation-significant} graph.
The models with \emph{+MP} contain the message-passing module. The models with \emph{+Dual} apply the dual encoder structure. 

\textbf{Effect of dual encoder structure.} We first validate the efficacy of applying dual structure to balance the importance of relations and objects by splitting our DM-GNN into two single models (\emph{Base-Obj} and \emph{Base-Rel}). Both single models perform poorly at $35.3\%\pm0.1\%$. This also shows that both relations and objects are vital to VQA performance. Absence of any of those modules leads to severe accuracy recession. 
Adding dual encoder structure~(\emph{+Dual}) leads to an empirical gain of \textbf{32.6\%} accuracy upward, which shows that the dual structure is significant in balancing relations and objects. 

\textbf{Effect of the message-passing module.} We validate the effectiveness of applying message-passing structure to learn a more comprehensive representation for scene graphs than the raw GGNN structure.
Comparing with the \emph{Base-Obj} and \emph{+MP} model, we note that after adding the message-passing structure, there is an improvement of 3.9\%. 
Comparing with the \emph{Base-Rel} and \emph{+MP} model, we observe an improvement of 3.6\%. 
The \emph{+Dual} model is DM-GNN without message-passing module. Comparing with DM-GNN, there is a 7.3\% improvement after adding the message-passing module. 
These empirical results show that message-passing structure can successfully improve the representation quality of scene graphs. 

\textbf{Effect of explicit modeling.} The \emph{w/o attr} model and the \emph{w/o rela} model remove the explicit attribute modeling and relation modeling part. Comparing \emph{w/o attr}, \emph{w/o rela} with DM-GNN, removing attribute modeling has a 3.6\% decrease in accuracy and removing relation modeling has 0.7\% decrease.

\vspace{-0.1in}
\subsection{Visualization}
\vspace{-0.1in}

\quad Fig.~\ref{visual}~(a) shows visualization on ``what" question type. Three ``what" question examples aimed at retrieving either object, relation or attribute information. 
Comparing row 1, row 2 with row 3, \emph{Obj} and \emph{Rel} models have strong attention bias toward objects and relations, while their combination \emph{Obj+Rel}, balances the attention on both sides and captures correct answers. 
Comparing row 3 with row 4, the message-passing module increases the score of correct answers. 

Fig.~\ref{visual}~(b) shows the visualization on ``why'', which need models to jointly exploit objects, relations and attributes to infer answers. With dual encoders and the message-passing module, our DM-GNN achieves \textbf{96.1\%}  on ``why" questions.


\vspace{-0.1in}
\section{Conclusion}
We propose DM-GNN, which encodes each scene graph into feature representations via an object encoder and a relation encoder generating a balanced and full-scale feature map using objects, attributes, and relations information, and demonstrate our model can effectively boost performances on GQA, VG and Motif-VG datasets . 


\bibliographystyle{IEEEbib}
\bibliography{real}

\end{document}


\sloppy

\def\x{{\mathbf x}}
\def\L{{\cal L}}

\title{AUTHOR GUIDELINES FOR ICME 2022 PROCEEDINGS}
%
\name{Anonymous ICME submission}
\address{}

\section{Scene graph generation}
For the scene graph generator(SGG), we follow the algorithms in ~\cite{tang2020sggcode}. First we'll introduce the evaluations metrics, then we'll introduce our algorithm's implementation details. This work is based on the deep learning framework PaddlePaddle\footnote{http://www.paddlepaddle.org/}.

\subsection{Evaluations}
We use \textbf{relationship retrieval(RR)} as our evaluation function. It can be further separated into three tasks: (i) Predicate Classification(\textbf{PredCls}): the model takes ground truth bounding boxes and labels as inputs. (ii) Scene Graph Classification(\textbf{SGCls}): the model uses ground truth bounding boxes without labels. (iii) Scene Graph Detection(\textbf{SFGet}): the model detects scene graphs from scratch. Relationship retrieval's conventional metric is \textbf{Recall@K(R@K)}. However, this conventional metric has severe bias \cite{DBLP:conf/cvpr/MisraZMG16}. To alleviate these problems, we use \textbf{mean Recall@K (mR@K)}~\cite{DBLP:conf/cvpr/ChenYCL19, DBLP:conf/cvpr/TangZWLL19} to replace the R@K metric. mR@K retrieves each predicate separately and then averages R@K for all predicates. 

\subsection{Implementation Details}
\subsubsection{Object Detector.} Following the previous works~\cite{DBLP:conf/cvpr/XuZCF17, DBLP:conf/cvpr/ZellersYTC18}, we use a frozed pre-trained Faster R-CNN~\cite{DBLP:conf/nips/RenHGS15} as the underlying detector of SGG models. We equipped the Faster R-CNN with a ResNet-101-FRN~\cite{DBLP:conf/cvpr/LinDGHHB17} backbone. The detector is trained on the VG training set using SGD as optimizer. The batch size is 8 and the initial learning rate to 8e-3. Four 2080ti GPUs are used for pretraining process.

\subsubsection{Generator Detials.} In scene graph generation part, on top of the frozen object detector, we train SGG models using SGD as optimizer. In PredCls and SGCls, the batch size is 12 and the initial learning rate is 12e-2. In SGDet, the batch size is 8 and the learning rate is 8e-2. The learning rate would be decayed by 10 two times after the validation performance plateaus. For SGDet, 80 RoIs were sampled for each image and Per-Class NMS~\cite{DBLP:journals/tc/RosenfeldT71} with 0.5 IoU was applied in object prediction. We sampled up to 1024 subject-object pairs containing 75\% background pairs during training. The performance is showed in Table~\ref{VG-detail}.

\section{DM-GNN Implementation Details}

\subsection{Datasets}

-- The \textbf{Visual Genome}~\cite{DBLP:journals/ijcv/KrishnaZGJHKCKL17} (VG) contains 108k fully annotated images. Compared with VG, \textbf{Motif-VG}~\cite{DBLP:conf/bmvc/ZhangCX19} has extra scene graph annotations with different qualities and biases. 

\noindent --  The \textbf{GQA} dataset~\cite{DBLP:conf/cvpr/HudsonM19} focuses on real-world reasoning and VQA tasks. It comprises 113k images and 22M questions.

We use 50-dimensional GLOVE word embeddings model ~\cite{pennington-etal-2014-glove} to embed words in the scene graph and questions. In order to record the questions' position information, we set up the positional encoding matrix $\textrm{PE}$:
\begin{gather}
    \textrm{PE}_{\texttt{pos}=2\emph{i}} = \sin(\texttt{pos}/10000^{2i/d_{\textrm{m}}}),\\
    \textrm{PE}_{\texttt{pos}=2\emph{i}+1} = \cos(\texttt{pos}/10000^{2i/d_{\textrm{m}}}) 
\end{gather}
where $\texttt{pos}$ is the position of the word in the question sequence. 
If $\texttt{pos}$ is odd, the position information is generated by a $\sin$ function, else, it is generated by a $\cos$ function. 
We also let model dimension equal to $d_{\textrm{m}}=50$.
After adding position information, the question embeddings are injected into a single-directional GRU network. 
The dimension of the hidden layers of the GRU is 100, and the dropout rate is 0.2.

In our message-passing enhanced GGNN encoder, the propagator time step is 5, and we use a bidirectional GRU as our message-passing module. 
Here, we set the dimension of the single GRU hidden layer to 50. 

In the fusion module, we apply a multi-head attention layer with 5 heads and no dropout. 
Regarding the answer predictor, we select the top-2000 answer candidates and use a 2-layer MLP as the output classifier.

We use Adam~\cite{kingma2014adam} as the optimizer, and Cross Entropy Loss as the loss function during the training of our model. 
For motif dataset, we set the batch size to 512. For Visual Genome ground truth dataset and GQA dataset, we set the batch size to 16 due to their abundant scene graph annotations.

The learning rate is decaying depending on the epoch number. We initialize the learning rate to be $1e^{-3}$, and when 30\% epochs finish, the learning rate drops to $2e^{-4}$. When 60\% epochs finish, the learning rate drops to $4e^{-5}$ and it becomes $8e^{-6}$ after 80\% epochs finish. We train our model and other baselines on a single V100 GPU.

\begin{table*}[ht]
\centering
    \resizebox{\textwidth}{15mm}{
    \begin{tabular}{l|l|l|lll|lll|lll}
    \hline
     Tasks &  &  &  & PredCls &  &  & SGCls &  &  & SFGet &  \\
    \hline
    \textbf{Model}&\textbf{Fusion}&\textbf{Method}&\textbf{mR@20}&\textbf{mR@50}&\textbf{mR100}&\textbf{mR@20}&\textbf{mR50}&\textbf{mR100}&\textbf{mR@20}&\textbf{mR@50}&\textbf{mR100}\\
    \hline
     IMP  & - & - & - & 9.8 & 10.5 & - & 5.8 & 6.0 & - & 3.8 & 4.8\\
     FREQ & - & - & 8.3 & 13.0 & 16.0 & 5.1 & 7.2 & 8.5 & 4.5 & 6.1 & 7.1\\
     VCTree & - & - & 14.0 & 17.9 & 19.4 & 8.2 & 10.1 & 10.8 & 5.2 & 6.9 & 8.0\\
     KERN & - & - & - & 17.7 & 19.2 & - & 9.4 & 10.0 & - & 6.4 & 7.3\\
    \hline
     MOTIF & SUM & baseline & 11.5 & 14.6 & 15.8 & 6.5 & 8.0 & 8.5 & 4.1 & 5.5 & 6.8\\
     MOTIF & SUM & Causal & 18.5 & 25.5 & 29.1 & 9.8 & 13.1 & 14.9 & 5.8 &8.2 & 9.8\\
    \hline
    \end{tabular}}
\caption{\label{VG-detail} The SGG performances of relationship retrieval on mean Recall@K. The MOTIF model is re-implemented under our codebase.}
\end{table*}

\begin{table*}
\centering
    \begin{tabular}{llllll}
    \hline
    \textbf{datasets/models}&\textbf{Motif} & \textbf{Causal-Motif}&\textbf{VCtree}&\textbf{IMP}&\textbf{GT}\\
    \hline
     NSM &43.1\% &42.9\%& 41.7\% & 43.9\% & 45.1\%\\
     FSTT & 48.1\%$\pm$0.02\% & 50.71\%$\pm$0.01\% & 52.85\%$\pm$0.05\% & 49.2\%$\pm$0.1\% & 67\%\\
     Re-GAT &54.5\%$\pm$1\% & 65.4\%$\pm$0.3\% & 71.6\%$\pm$0.2\% &44.7\%$\pm$0.7\% &71\%\\
     DM-GNN (Ours) & \textbf{72.9\%}$\pm$0.4\%& \textbf{73.2\%}$\pm$0.3\% & \textbf{74.3\%}$\pm$0.5\% & \textbf{71.3\%}$\pm$0.1\% &\textbf{76\%}\\
    \hline
    \end{tabular}
\caption{\label{citation-guide}
Performance on VG test dataset with different scene graph reasoning methods. Consistent improvements are observed and it demonstrates that our model outperforms scene graph reasoning models for VQA.}
\end{table*}

\begin{table*}[htbp]
    \begin{floatrow}
    \capbtabbox{
     \begin{tabular}{lllll}
    \hline
    \textbf{Models}&\textbf{NSM} & \textbf{FSTT}&\textbf{ReGAT}&\textbf{DM-GNN}\\
    \hline
     Acc.(\%) & 35.5 & 31.6 & 54.5 & \textbf{54.9} \\
    \hline
    \end{tabular}
    }{
     \caption{Model accuracy on Motif dataset without question fusion module.}
     \label{model-without-fusion}
    }
    \capbtabbox{
        \begin{tabular}{llll|l}
        \hline
         \textbf{Models} & FSTT & ReGAT & DM-GNN & Total \\
        \hline
         Relation & 45.9\% & 44.6\% & \textbf{32.9}\% & 7913\\
         Object &58.6\% & 47.3\% & \textbf{25.2}\% & 4414\\
         Attribute & 49.6\% & 22.8\% & \textbf{16.8}\% & 15483\\
        \hline
        \end{tabular}
    }{
     \caption{Error rate analysis on motif-VG dataset.}
     \label{badcase}
    }
    \end{floatrow}
\end{table*}

\section{More Experiments}
Table~\ref{citation-guide} reports results on the test sets of four different datasets generated from the VG dataset. Compared to the baseline models, we can observe that our DM-GNN model outperforms the other baselines. In Motif, Unbias-Causal, and IMP datasets, our model has a 10\% to 20\% improvement in performance.

In addition, the stability of our model is conspicuous. Our model can achieve consistent and stable performance under different scene graph qualities. Our model's performances under four different scene graph datasets only suffer 3\% fluctuation,
while FSTT suffers 4.7\% fluctuation and Re-GAT suffers 26.9\% fluctuation. Our model can also steadily converge at nearly 15 epochs under four different datasets with various scene graph qualities, which is faster than others.

The question fusion module, which concats the question vector with the reasoning vector before entering into the answer predictor module, commonly appears in VQA models, including NSM, FSTT, and our DM-GNN model. This method can indeed improve the accuracy of VQA models. However, from the cognitive aspect, the question fusion module lacks interpretability because the reasoning vector in the model has included question features. Also, the addition of the question fusion module may lead the reasoning model to "guess answers" from questions, which negatively influences reasoning itself. We retrain our model and other baselines without the question fusion module to evaluate the reasoning ability without the influence of outer question information. Table~\ref{model-without-fusion} shows the results of models without question fusion module. Noted that there is no question fusion module in Re-GAT baseline, so the Re-GAT result is the same as Table~\ref{citation-guide}.

\textbf{Error Rate Analysis} To demonstrate that our dual encoders structure can intensify the model's perception of relation features and learn a comprehensive representation from nodes, attributes, and relations information, we establish an error rate analysis for baselines and DM-GNN on motif-VG.

The badcases are classified into object, relation and attribute. We present in Table~\ref{badcase} the results for our error rate analysis. Our DM-GNN model surpasses all baselines in the terms of objects detection and our model does well in relation retrieval, outperforming GNN based FSTT and Re-GAT. This proves our model does alleviate the unbalance focus on objects and relations.
Also, our model reduces nearly half of the wrong answers in FSTT, Re-GAT in the attribute aspect, which greatly improves the false attribute selection phenomenon.

\subsection{More Visualization}

\quad Fig.~\ref{visualbis} exhibits the details of top 5 attention scores for different question types. Our dual structures and message-passing module significantly increase the attention scores of correct answers.

To better illustrate the effectiveness of the dual encoder structure and the message-passing module in our DM-GNN model, we compare the attention scores learned by DM-GNN model with those learned by our baseline, object-MP, and relation-MP models. Figure~\ref{visualize} is the detail of the visualization result. Row 1 is three typical images with questions. 

The attention scores are calculated from the similarity between the total feature map and the question encoder vector.

\begin{figure*}[ht] 
    \centering 
    \includegraphics[width=1\textwidth]{./pic/visual_detail.pdf} 
    \caption{Examples of top 5 attention scores of candidate answers for different types of questions.} 
    \label{visualbis} 
\end{figure*}

\begin{figure*}
    \centering
    \includegraphics[scale=0.43]{./pic/visual.pdf}
    \caption{More visualization for our DM-GNN model.}
    \label{visualize}
\end{figure*}

\clearpage

\bibliographystyle{IEEEbib}
\bibliography{icme2022template}